%% file: cdmamba.tex
\documentclass[10pt,twocolumn,letterpaper]{article}
\usepackage[utf8]{inputenc}
\usepackage[T1]{fontenc}
\usepackage{iccv}              


\definecolor{iccvblue}{rgb}{0.21,0.49,0.74}
\usepackage[pagebackref,breaklinks,colorlinks,allcolors=iccvblue]{hyperref}

\usepackage{caption}
\usepackage{subcaption}
\usepackage{graphicx}
\usepackage[american]{babel}
\usepackage{microtype}
\usepackage{multirow}
\usepackage{booktabs}
\usepackage{algorithm}
\usepackage{algorithmicx}
\usepackage{algpseudocode}

\usepackage{bm}

\usepackage{wrapfig}
\usepackage{pifont}
\usepackage{multirow}
\usepackage[export]{adjustbox}
\usepackage{color}
\usepackage{colortbl}
\usepackage{lipsum}
\usepackage{pifont}
\usepackage{bbding}
\usepackage[dvipsnames]{xcolor}
\usepackage{array}

\usepackage[capitalize]{cleveref}
\usepackage{float}
\crefname{section}{Sec.}{Secs.}
\Crefname{section}{Section}{Sections}
\Crefname{table}{Table}{Tables}
\crefname{table}{Tab.}{Tabs.}
\definecolor{red}{RGB}{255,0,0}
\definecolor{blue}{RGB}{0,0,255}
\definecolor{green}{RGB}{0,255,0}
\definecolor{mygray}{gray}{.9}
\definecolor{mygray2}{gray}{.5}
\definecolor{mywarning}{RGB}{233,144,61}
\definecolor{mygreen}{RGB}{93,174,86}
\definecolor{codefunc}{RGB}{73,122,234}

\usepackage[american]{babel}
\usepackage{microtype}

\definecolor{mygreen}{RGB}{0,154,85}
\definecolor{myy}{RGB}{126,95,0}
\definecolor{myred}{RGB}{212,121,116}
\definecolor{myblue}{RGB}{184, 134, 73}
\definecolor{mynewgreen}{RGB}{113,188,169}
\definecolor{mypurple}{RGB}{123,104,238}
\colorlet{R1}{myblue}
\colorlet{R2}{mypurple}
\colorlet{R3}{myred}
\colorlet{R6}{mypurple}
\definecolor{mycite}{RGB}{73,123,184}
\colorlet{cite}{mycite}

\newcommand{\B}{{\bm{b}}}

\newcommand{\h}{{\bm h}}

\newcommand{\D}{{\rm d}}
\newcommand{\T}{{\top}}

\newcommand{\F}{{\bm F}}

\def\paperID{1234} 
\def\confName{ICCV}
\def\confYear{2030}

\title{CD-Mamba: Cloud Detection with Long-Range Spatial Dependency Modeling}

\author{
	Tianxiang Xue$^1$, Jiayi Zhao$^1$, Jingsheng Li$^1$, Changlu Chen$^2$, and Kun Zhan$^{1,\star}$\\
	1. School of Information Science and Engineering, Lanzhou University\\
	2. Faculty of Data Science, City University of Macau, Macau, China\\
	{\small \url{https://github.com/kunzhan/CD-Mamba}}
}

\begin{document}
\maketitle
\input{body}

{
\small
\bibliographystyle{ieeenat_fullname}
\bibliography{tgrs}
}
\end{document}

%% file: body.tex
\begin{abstract}
Remote sensing images are frequently obscured by cloud cover, posing significant challenges to data integrity and reliability. Effective cloud detection requires addressing both short-range spatial redundancies and long-range atmospheric similarities among cloud patches. Convolutional neural networks are effective at capturing local spatial dependencies, while Mamba has strong capabilities in modeling long-range dependencies. To fully leverage both local spatial relations and long-range dependencies, we propose CD-Mamba, a hybrid model that integrates convolution and Mamba's state-space modeling into a unified cloud detection network. CD-Mamba is designed to capture pixel-wise textural details and long-term patch-wise dependencies for cloud detection. This design enables CD-Mamba to manage both pixel-wise interactions and extensive patch-wise dependencies simultaneously, improving detection accuracy across diverse spatial scales. Extensive experiments validate the effectiveness of CD-Mamba and demonstrate its superior performance over existing methods.
\end{abstract}

\begin{figure}[!t]
	\centering
	\includegraphics[width=0.98\linewidth]{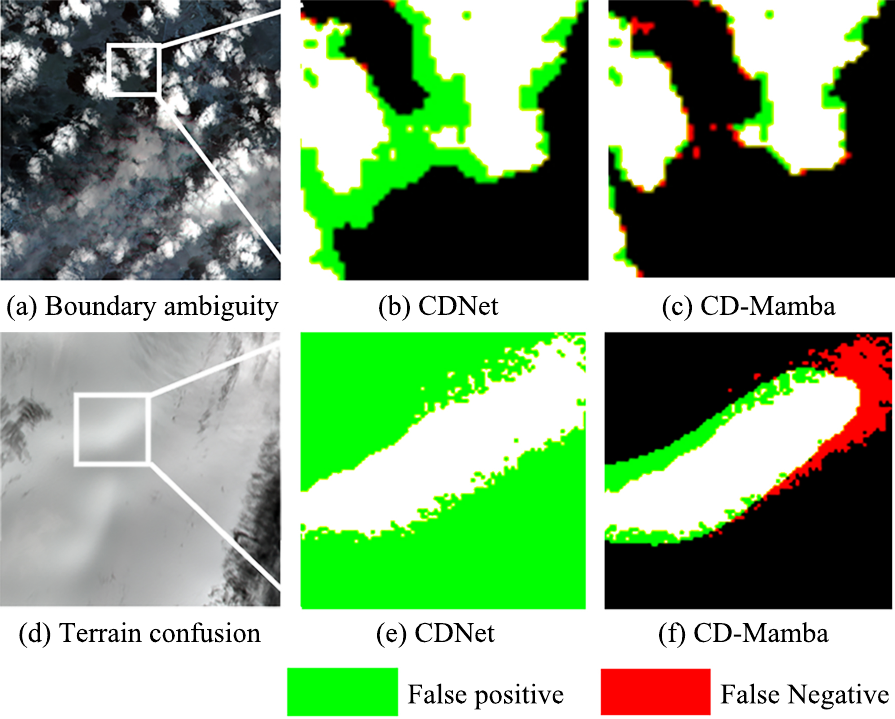}
	\caption{
		Two classic challenges in cloud detection are boundary ambiguity and terrain confusion. (a) depicts a cloud image with boundary ambiguity, while (d) shows a cloud-snow mixed terrain. (b) and (e) present segmentation results from CDNet, and (c) and (f) show results from CD-Mamba. The comparison highlights CD-Mamba’s ability to address two critical challenges: (1) mitigating boundary ambiguity and (2) CD-Mamba reduces false positives (e.g., snow-to-cloud confusions), outperforming traditional methods like CDNet in complex scenarios. The green areas indicate false positives (incorrectly identified as clouds), and the red areas represent false negatives (actual clouds missed by the detection). The white regions, where predictions overlap with the ground-truth labels, signify correct detections.
	}\label{show}
\end{figure}
\section{Introduction}
Approximately 66\% of Earth's surface is often obscured by clouds~\cite{zhang2004calculation, king2013spatial}, which can significantly hinder satellite observations by blocking critical surface features. This widespread cloud cover presents a major challenge for remote sensing, where clear views of Earth's surface are essential for accurate data collection. Consequently, effective cloud detection methods are crucial for ensuring the integrity and reliability.

Over the past decade, convolutional neural networks (CNNs)~\cite{li2018cloud, yang2019cdnet, rs13224533} and vision transformers (ViTs)~\cite{zhang2022cloudvit, 10171167, hu2023mcanet} have been widely applied to cloud detection. CNNs are good at capturing local spatial dependencies due to the small receptive fields of convolutional kernels~\cite{he2016deep}, making them effective at identifying cloud textural features. Meanwhile, ViTs, with their patch-based tokenization~\cite{dosovitskiy2020vit}, capture broader spatial patterns by modeling relationships between image patches. While CNNs focus on pixel-level relationships, ViTs model patch-wise associations, offering an advantage in detecting spatially discontinuous clouds. Mamba~\cite{gu2023mamba, guefficiently} further enhances this capability by modeling long-range dependencies between tokens.

Recently, vision Mamba~\cite{zhu2024vision,yang2024vivim,liu2024vmamba,liu2024swin,liu2024visionmamba,Yu_2025_CVPR,maxjars2025} has demonstrated significant potential in modeling long-range dependencies between patches, emerging as a promising approach that complements CNNs. While CNNs are well-suited for capturing short-range local spatial correlations, Mamba addresses their limitation by efficiently processing global contextual relationships. This capability is particularly advantageous for cloud detection, where clouds often span extensive and discontinuous regions.

In existing cloud detection methods, two major challenges limit accuracy: boundary ambiguity~\cite{acp-16-3463-2016} and terrain confusion~\cite{shao2019cloud}. As illustrated in Figure~\ref{show}(a), boundary ambiguity arises when cloud edges blend with the background, making it difficult to delineate cloud contours precisely. On the other hand, terrain confusion, shown in Figure~\ref{show}(d), occurs when clouds exhibit similar spectral characteristics to snow-covered or bright terrain, leading to misclassification. Various approaches have been proposed to address the two challenges in cloud detection. KET-CD~\cite{kcd} alleviates the terrain confusion that occurs during model learning by introducing a novel label-aware embedding mechanism. DB-Net~\cite{dbnet} improves the model’s ability to detect small and thin cloud boundaries by combining the strengths of  CNNs and Transformers. DABNet~\cite{9314019} enhances multi-scale texture modeling through a pyramid architecture and incorporates a weighted loss function to better focus on cloud boundaries. However, these methods typically target only one specific issue and fail to address the two core challenges in cloud detection: boundary ambiguity and terrain confusion, thereby limiting overall performance improvement.

To tackle these challenges, we propose CD-Mamba, which integrates the global contextual reasoning capability of Mamba with the local detail extraction strength of CNNs. This hybrid design enables the model to effectively distinguish between spatially fragmented yet semantically connected cloud regions shown in Figure~\ref{show}(c) and to suppress false positives caused by terrain-like features shown in Figure~\ref{show}(f). Cloud images typically exhibit both fine-grained textures and long-range spatial dependencies. Accurate boundary delineation often requires understanding the broader spatial context—especially when a single cloud spans disjoint regions. As shown in Figure~\ref{show}(a), these nonadjacent cloud segments necessitate holistic modeling of spatial dependencies. In complex remote sensing scenarios, such as snow-covered landscapes where clouds and background share similar visual characteristics shown in Figure~\ref{show}(d), incorporating long-range spatial relationships becomes crucial. By modeling these relationships explicitly, CD-Mamba significantly improves both the accuracy and robustness of cloud detection, especially in challenging conditions involving boundary ambiguity and terrain confusion. By combining convolutional blocks for local representation, spatial Mamba blocks for global dependency modeling, and dual-attentional skip connections for refined feature fusion, CD-Mamba forms a unified architecture tailored for complex cloud detection tasks. This hybrid design substantially improves segmentation accuracy, particularly in challenging regions with boundary ambiguity or terrain confusion.

To preserve high-resolution spatial features and facilitate multi-scale feature fusion, CD-Mamba adopts a U-shaped architecture inspired by U-Net~\cite{ronneberger2015u}. The skip connections between the encoder and decoder are enhanced with dual attention modules, consisting of spatial and channel-wise attention mechanisms~\cite{sun2023datransunet,ruan2022malunet}. These modules highlight informative regions and reduce redundancy, leading to more effective integration of shallow and deep features. CD-Mamba is designed to simultaneously capture fine-grained local details and long-range contextual dependencies, both of which are essential for accurate cloud detection. At the initial stage, a convolutional block is used to extract local spatial features and fine-grained textures, which are crucial for delineating precise cloud boundaries. To capture broader spatial context, we introduce a cloud spatial-Mamba block (Cloud-SMB), which efficiently models long-range dependencies. Cloud-SMB adopts a multi-directional scanning strategy, allowing the model to learn spatial relationships from various directions. To further improve efficiency, we employ a parallel processing strategy that splits the feature maps along the channel dimension while sharing model parameters across branches. This design is particularly advantageous for handling large, fragmented cloud structures in remote sensing imagery.

In summary, the main contributions of the paper are:
\begin{itemize}
\item The paper introduces CD-Mamba, an architecture that enables the model to capture both fine-grained spatial details and broader contextual patterns, significantly enhancing the accuracy of cloud detection.
\item  CD-Mamba effectively integrates the strengths of U-shaped networks and feature fusion through skip connections, enabling the model to emphasize key spatial regions and channel-specific features, which improves the integration of multi-scale information.
\item  CD-Mamba addresses two challenges in cloud detection: boundary ambiguity and terrain confusion, improving performance in complex remote sensing scenarios.
\end{itemize}
\section{Related works}

\subsection{Local and Short-Range Dependency Modeling.}
Recent deep learning models have demonstrated strong capabilities in cloud detection and remain widely used in this field~\cite{li2018cloud, li2022cloud}. For example, CDNet~\cite{yang2019cdnet} enhances the expressive power of convolutional
networks by incorporating a deeper ResNet~\cite{he2016deep}. As shown in Figure
\ref{show}, CDNet struggles in complex scenarios such as ambiguous cloud boundaries and terrain confusion, where accurate discrimination remains challenging. CDnetV2~\cite{guo2020cdnetv2} further improves upon CDNet by introducing channel and spatial attention-based feature map fusion and high-level fusion feature extraction, fully leveraging the features extracted from encoder layers to produce accurate cloud detection results. DABNet~\cite{9314019} proposes a
deformable contextual feature pyramid module to enhance multi-scale feature modeling and employs
a boundary-weighted loss function to focus on cloud boundary information. HR-cloud-Net~\cite{10.1117} adopts a hierarchical high-resolution integration approach, effectively capturing complex cloud
features while facilitating feature exchange across different resolutions, thereby improving the
overall performance of cloud detection.

The advent of ViTs has made it possible for cloud detection to capture longer spatial dependencies. For example, CloudViT~\cite{zhang2022cloudvit} uses a transformer as the network backbone and incorporates the dark channel prior to guide feature learning. Subsequently, an increasing number of researchers have adopted hybrid architectures combining CNNs and ViTs to leverage the strengths of both technologies. CNNs are renowned for their powerful local feature extraction capabilities. Their convolutional layers can effectively capture fine-grained details such as the edges and textures of cloud layers. On the other hand, the self-attention mechanism of ViTs can capture the broader range dependencies across large areas in cloud images, enabling the model to more accurately identify the boundaries of cloud layers and distinguish between different types of clouds, thus significantly enhancing the accuracy and integrity of cloud detection. For instance, CNN-TransNet \cite{10171167} integrates the advantages of ViTs and CNNs, enhancing detail handling and the establishment of relatively broader range dependencies to process rich cloud layer information. MCANet~\cite{hu2023mcanet} also combines the strengths of CNNs and ViTs, constructing a multi-branch feature extraction structure that addresses the issue of missing spatial information in large-scale remote sensing images.

Despite the progress achieved by deep learning in cloud detection, several challenges remain. These approaches primarily focus on limited-range spatial dependencies, which may constrain their effectiveness in capturing global contextual features, especially in complex or large-scale cloud scenarios. Moreover, the high computational cost and hardware demands of ViTs, combined with the multispectral and high-resolution characteristics of remote sensing images, further constrain their practical applicability.
\subsection{Long-Range Dependency Modeling.}
Cloud features possess rich spatial information. Vision Mamba models can effectively integrate the global contextual information of remote sensing images, enhancing the understanding and representation of complex cloud boundaries~\cite{zhu2024vision,yang2024vivim,liu2024vmamba,liu2024swin,liu2024visionmamba,Yu_2025_CVPR,maxjars2025}. Therefore, we have pioneered the introduction of the Vision Mamba models network into cloud detection tasks. Mamba captures global modeling capabilities while achieving linear computational complexity through hardware-optimized parallel scanning algorithms. This avoids the high computational complexity of ViTs when processing long sequences. Its excellent performance has been demonstrated across a variety of visual tasks. For example, VM-UNet~\cite{ruan2024vm} combines the Mamba model with a symmetric encoder-decoder structure to process long-sequence data in biomedical image segmentation, enabling more comprehensive extraction of biological information. Zhu et al.~\cite{zhu2024vision} designs a bidirectional Mamba module to construct a novel Mamba backbone network, excelling in tasks such as image classification and semantic segmentation. UltraLight VM-UNet~\cite{wu2024ultralight} focuses on exploring lightweight Mamba models while retaining the symmetric structure of U-Net and introducing skip connections, significantly improving the accuracy of semantic segmentation tasks. RS-Mamba~\cite{zhao2024rs} leverages its linear complexity and global modeling capabilities to process large-scale remote sensing images, particularly excelling in dense prediction tasks for ultra-high-resolution remote sensing imagery.

To better balance local and global dependencies in cloud detection, CD-Mamba combines convolutional layers with Mamba's long-range modeling. This hybrid design allows the model to capture both fine-grained textures and global cloud structures, enhancing detection performance across diverse spatial patterns.
\section{CD-Mamba for Cloud Detection}
The objective is to train a neural network $Y= \sigma(X|\theta)$ that maps an input multispectral satellite image $X\in\mathbb{R}^{\lambda \times h \times w}$ into a pixel-wise probability $Y\in[0, 1]^{h \times w}$ via a sigmoid activation, where $\lambda$ denotes the number of spectral channels, $h$ and $w$ represent the image height and width, respectively. The parameter $\theta$ refers to the model parameters that are learned during training. Each value in $Y$ represents the estimated likelihood that the corresponding pixel belongs to a cloud region.

\begin{figure*}[!t]
\centering %
\includegraphics[width=0.95\textwidth]{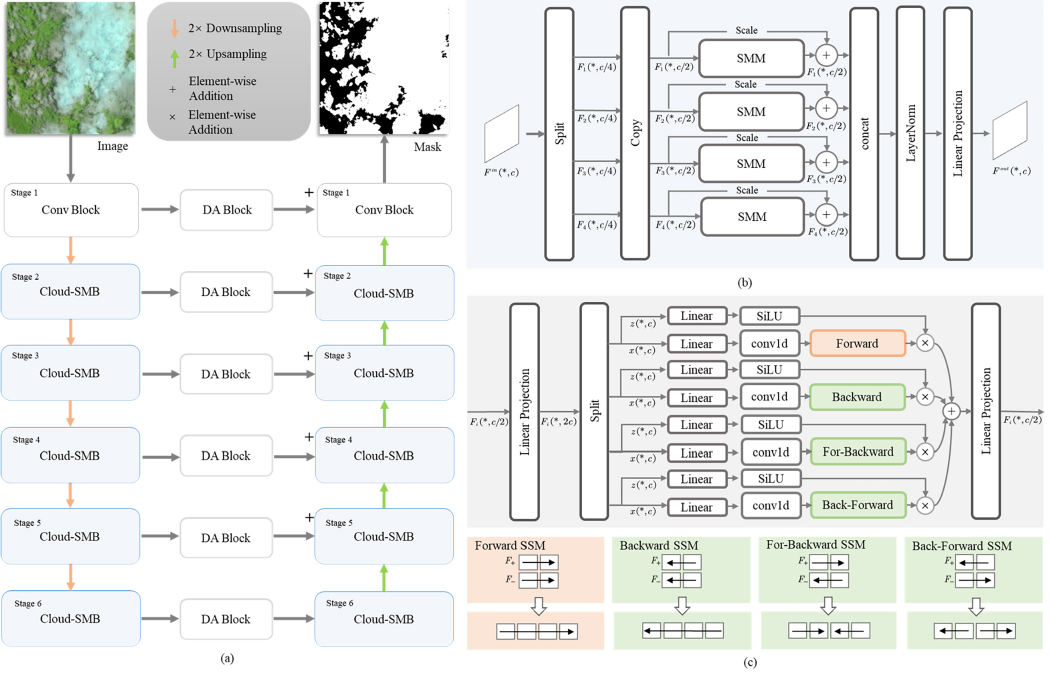}
\caption{CD-Mamba overview. CD-Mamba primarily integrates convolutional blocks and the Cloud-SMB hybrid structure within a U-shaped network architecture. The skip connections are enhanced with dual-attention modules. (a) shows the overall CD-Mamba network design, (b) illustrates a single Cloud-SMB, which contains four parallel, parameter-shared SMMs, and (c) demonstrates an SMM within the Cloud-SMB. An SMM consists of four parallel SSMs and employs a multi-directional scanning strategy. We distinguish the SSM modules that process the backward, forward-backward, and backward-forward directional sequences by using a different color (green) from the forward direction (red), to indicate that parameter sharing is employed in the former.}
\label{architecture}
\end{figure*}

We introduce CD-Mamba as shown in Figure~\ref{architecture}(a), a novel approach for cloud detection. CD-Mamba combines the strengths of CNNs for capturing local spatial dependencies and the cloud spatial-Mamba block hybrid structure for modeling long-range contextual relationships. By integrating these components within a U-shaped network architecture, CD-Mamba leverages both local and global information. Additionally, dual-attention modules in the skip connections further enhance the feature fusion process.
\subsection{State-Space Model.}
A continuous-time state-space model~\cite{kalman1960new} is defined by the following two equations:
\begin{align}
\frac{\D \h(t)}{\D t}
&= A\h(t)+\bm{b}x(t) \,, \\
y(t)
&= \bm{c}^\T h(t) \,,
\end{align}
where $\h(t)\in \mathbb{R}^n$ is the state vector at time $t$, $x(t)\in \mathbb{R}$ is the input, $y(t)\in \mathbb{R}$ is the observed output, $A\in\mathbb R^{n\times n}$ is the state transition matrix, and $\bm b$ and $\bm c$ are system parameters with appropriate dimensions matching $\h(t)$.

To discretize the continuous-time state-space model, zero-order holding is applied to the inputs~\cite{gu2023mamba, guefficiently}, resulting in the following discrete-time equations:
\begin{align}
\h_i
 &=\bar{A}\h_{i-1}+\bar{\bm{b}}x_i \,, \\
y_i
&=\bm{c}^\top\h_i \,,
\end{align}
where $\bar{A}=e^{\Delta A}$ is the discretized state transition matrix, $\bar{\B}=(\Delta A)^{-1}(e^{\Delta A}-I)\Delta\B$ is the discretized input matrix, $\Delta$ is the sampling period between discrete time steps $i-1$ and $i$, and $I$ is the identity matrix.

By iterating the discrete-time state-space equations starting from the first time step, we can derive the output at time step $l$ as a function of the input sequence $\boldsymbol{x} = [x_1, x_2, \ldots, x_l]^\top$. This results in the following expression:
\begin{align}
y_l = [\bm{c}^\top \bar{A}^l \bar{\bm{b}}, \ldots, \bm{c}^\top \bar{A}^2 \bar{\bm{b}}, \bm{c}^\top \bar{A} \bar{\bm{b}}] \boldsymbol{x} \,,
\end{align}
where the output $y_l$ is represented as a linear combination of the past inputs, and each coefficient captures the influence of the corresponding input on the current output through repeated applications of the state transition matrix $\bar{A}$. By recording the outputs at each time step, we obtain the output sequence $\boldsymbol{y} = [y_1, y_2, \ldots, y_l]^\top$.

In the context of the vision Mamba approach, the parameters $\bar{A}$, $\bar{\bm{b}}$, $\bm{c}$, and $\Delta$ undergo specific modifications~\cite{zhu2024vision,yang2024vivim,liu2024vmamba,liu2024swin,liu2024visionmamba,Yu_2025_CVPR,maxjars2025}. The state transition matrix $\bm{A}$ is redefined as a learnable parameter with dimensions $c \times n$, where $c$ represents the feature dimension and $n$ represents the hidden state dimension. The parameters $\bm{B}$ and $\bm{C}$ are extended to account for the sequence length $l$ and mini-batch size $b$, with dimensions $b \times l \times n$, mapping the input sequence to the hidden states. $\bm{\Delta}$ is redefined as a learnable parameter with dimensions $b \times l \times c$, capturing the temporal dynamics and feature dependencies over time.

A selective state-space model (SSM) is then defined by $\bm{F}_j = \text{SSM}(\bm{F}_i, \bm{A}(\theta), \bm{B}, \bm{C}, \bm{\Delta}(\theta))$, where both $\bm{F}_i$ and $\bm{F}_j$ have dimensions $b \times l \times c$. In this equation, $\bm{F}_i$ and $\bm{F}_j$ represent the inputs and outputs of the SSM, respectively.

\begin{algorithm}[!t]
\caption{The Cloud-SMB algorithm.}\label{algo}
\begin{algorithmic}[1]
\Require $\F^{\rm in}$\,; \Comment{$(b, l, c)$}
\Ensure  $\F^{\rm out}$\,; \Comment{$(b,l, c)$}
\State $\F =  \text{norm}{(\F^{\rm in})}$\,; \Comment{LayerNorm}
\State $\F_1,\F_2,\F_3,\F_4 = \text{split}(\F, c)$\,; \Comment{$\forall\,i,\F_i$: ($b,l,$$\frac c4$)}
\For{$i\in[1,4]$}
\State $\F'_i=\F_i$\,; \Comment{Copy}
\State $\F_f=[\F_i,\F'_i]$\,;\Comment{Forward}
\State $\F_{b} = \text{flip}{(\F_i)}$ \,;  \Comment{Backward}
\State $\F_{+},\F_{-}  = \text{split}(\F_i, l)$\,;
\State $\F_{fb} = [\F_+,\text{flip}{(\F_-)}]$\,;\Comment{For-backward}
\State $\F_{bf} = [\text{flip}{(\F_+)}, \F_-]$\,;\Comment{Back-forward}
\For{$o\in\{{f}, {b}, {fb}, {bf}\}$}
\State $\F_o = \text{SSM}(\F_o)$ \,;
\EndFor
\State $\F_i = \text{linear}(\F_f + \F_b +\F_{fb}+\F_{bf}) + \F_i$\,;
\EndFor
\State $\F= [\F_1,\F_2,\F_3,\F_4]$\,; \Comment{$(b,l, 2c)$}
\State $\F = \text{norm} (\F)$\,;
\State \Return $\F^{\rm out} = \text{linear} (\F)$\,;  \Comment{$(b,l, c)$}
\end{algorithmic}
\end{algorithm}

\subsection{Cloud-SMB.}
As shown in Figure~\ref{architecture}(b), a Cloud-SMB consists of four SMMs. As shown in Figure~\ref{architecture}(c), each SMM performs scanning in four directions. The pseudocode for the feature processing procedure of a Cloud-SMB is listed in Algorithm~\ref{algo}.

We propose a parallel processing strategy for Cloud-SMB, as shown in Figure~\ref{architecture}(b). The input feature map $\F^{\text{in}}$ is first normalized and then evenly divided into four feature maps $\F_i$ along the channel dimension~\cite{wu2024ultralight}, each with a dimensionality of $\frac{c}{4}$. To preserve the input relevance within the gating mechanism of the Mamba update, a copy of each feature map is created, denoted as $\F'_i$. Each $\F_i$ is concatenated with its corresponding copy $\F'_i$ along the channel dimension,  resulting in intermediate features of dimensionality $\frac{c}{2}$. The above feature processing procedure is detailed in lines 1–5 of Algorithm~\ref{algo}.

To effectively extract spatial information, we design a four-directional scanning strategy, as illustrated in Figure~\ref{architecture}(c). The forward sequence $\F_f$ is obtained by scanning the original input sequence $\F_i$ in its natural order, while the backward sequence $\F_b$ is generated by reversing $\F_i$ along the sequence dimension. To construct more diverse directional contexts, $\F_i$ is further divided into two equal parts, $\F_+$ and $\F_-$, representing the first and second halves of the sequence, respectively. The for-backward sequence $\F_{fb}$ is formed by concatenating $\F_+$ with the reversed $\F_-$, while the back-forward sequence $\F_{bf}$ is constructed by concatenating the reversed $\F_+$ with $\F_-$. The four-directional scanning mechanism enhances the model's ability to capture diverse spatial features in images and significantly improves cloud detection performance. For implementation details, please refer to lines 6–10 of Algorithm~\ref{algo}.

Subsequently, the sequences obtained from the four directions are updated by SSM. These updated sequences are combined and linearly mapped to aggregate features, which are then added to the original features $\F_i$ to generate the corresponding outputs. Finally, we join and normalize the outputs of all feature maps $\F_i$ along the channel dimension, and reduce the number of channel to $c$ using a linear mapping to obtain the final output feature graph $\F^ {\rm out} $.
\subsection{The CD-Mamba Algorithm.}
As shown in Figure~\ref{architecture}(a), CD-Mamba consists of three main components: an encoder, a decoder, and skip connection modules. Both the encoder and decoder are divided into six stages, containing a convolutional block and five Cloud-SMBs.

The input image is initially processed by a $3 \times 3$ convolutional layer in Stage 1 of the encoder, with both stride and padding set to 1. This is followed by five consecutive Cloud-SMB stages.

Since the five Cloud-SMB stages share a similar structure, we use Stage 2 as an example to illustrate the feature processing workflow. First, the output feature map $\F_1^{\rm enc}$ from Stage 1 is normalized using group normalization~\cite{abs-1803-08494}. It is then downsampled via a $2 \times 2$ max-pooling operation, followed by activation through the Gaussian Error Linear Unit (GELU)~\cite{DBLP:HendrycksG16}. These steps are collectively represented by the orange arrows in Figure~~\ref{architecture}(a). The processed feature map $\F_2^{\rm enc}$ is then obtained at the end of Stage 2.

The subsequent stages adopt the same configuration as Stage 2, ultimately producing multi-scale feature representations $\F_i^{\rm enc}$, which encode the input image at different spatial resolutions. To facilitate effective feature aggregation, skip connections are introduced. Specifically, the input feature maps are processed through skip connections to yield $\F_i^{\rm skip}$, which preserve the spatial scale and channel dimensions consistent with their corresponding encoder outputs.

In the decoder of CD-Mamba, the spatial resolution is progressively restored through a series of upsampling operations between blocks. After passing through five Cloud-SMB blocks, the final feature map is processed by a convolutional block to generate the cloud probability map $Y$.

As shown in Figure~\ref{architecture}(a), Stage 6 of the encoder is connected to Stage 6 of the decoder. The encoder output $ \F_6^{\rm enc} $ is first passed through a skip connection to produce $ \F_6^{\rm skip} $, which is then fed into the decoder Stage 6 to produce $ \F_5^{\rm dec} $. Skip connections are used to fuse encoder features at corresponding scales, thereby enhancing the decoder's ability to recover fine-grained spatial details.

The subsequent Stages 5, 4, 3, and 2 of the decoder follow a similar design. We take Stage 5 as an illustrative example. At Stage 5, the input consists of two components $ \F_5^{\rm dec} $ and $ \F_5^{\rm skip} $. The two features are fused via element-wise addition. The fused feature then undergoes processing in Stage 5, including group normalization, $ 2 \times 2 $ bilinear upsampling, and GELU activation, ultimately producing $ \F_4^{\rm dec} $.

$\F_4^{\rm dec}$ and $\F_4^{\rm skip}$ are subsequently fed into Stage 4 of the decoder. This process is repeated for the remaining Cloud-SMB Stages of the decoder. In Stage 1 of the decoder, to better capture sharp and accurate cloud boundaries, a convolutional block is employed,  followed by a sigmoid activation to produce the cloud probability map $Y$.

To optimize cloud detection performance, we adopt a composite loss function that combines binary cross-entropy and Dice loss~\cite{dice}. Although the network only outputs the cloud probability, the task itself is a binary classification between cloud and background. Therefore, the binary cross-entropy loss is used to provide strong supervision for both cloud and background, ensuring pixel-level classification consistency. The binary cross-entropy loss is given by
\begin{align}
\mathcal L_{\rm bce} = -\frac{1}{hw} \sum_{i=1}^{hw} \left[ t_i \log(y_i) + (1 - t_i) \log(1 - y_i) \right],
\end{align}
where $y_i$ is the pixel-wise cloud predicted probability, $t_i$ is the ground-truth label, and $hw$ is the total number of pixels.

However, due to the inherent imbalance between the number of cloud and background pixels, we also incorporate the Dice loss, which is more robust to class imbalance and helps improve the overlap between predictions and ground truth, especially for sparse cloud regions. The Dice loss~\cite{dice}, focusing on region-level similarity, is formulated as:
\begin{align}
\mathcal L_{\rm dice} = 1 - \frac{2\sum_{i=1}^{hw} y_i t_i + \varepsilon}{\sum_{i=1}^{hw} y_i + \sum_{i=1}^{hw} t_i + \varepsilon} \,,
\end{align}
where $\varepsilon$ is a smoothing factor to avoid division by zero.

The overall loss is a weighted sum of the two terms:
\begin{align}
\mathcal L_{\rm overall} = \mathcal L_{\rm bce} + \gamma \mathcal L_{\rm dice} \,,
\end{align}
where $\gamma$ is a trade-off parameter.
\subsection{Skip Connection of CD-Mamba.}
Skip connections are essential in U-shaped architectures for binary discriminative tasks, as they facilitate spatial detail recovery and enhance training stability. We incorporate a Dual Attention Block (DA-Block) into the skip connections, serving as a feature enhancement module. This design significantly improves the efficiency of internal information flow~\cite{Woo_2018_ECCV}.

DA-Block takes the encoder output $\F_i^{\rm enc}$ from the corresponding stage as input and generates $\F_i^{\rm skip}$ as the enhanced skip connection feature. The computation of DA-Block is formulated as follows:
\begin{align}
\F_i^{\rm pam}
&=\text{PAM}(\F_i^{\rm enc})+\F_i^{\rm enc} \,,\\
\F_i^{\rm skip}
&= \text{CAM}(\F_i^{\rm pam})+\F_i^{\rm pam}+\F_i^{\rm enc} \,,
\end{align}
where PAM represents the position attention module and CAM denotes the channel attention module.

For the attention modules within the skip connections of CD-Mamba, Stages 4-6 adopt the design from Sun \textit{et al.}~\cite{sun2023datransunet}, while Stages 1-3 follow the structure proposed in Ruan \textit{et al.}~\cite{ruan2022malunet}.

Depending on the feature map resolutions at different stages of the network, we implement two distinct attention mechanisms: PAM and CAM. In Stages 4 to 6 of CD-Mamba, we utilize a key-value attention mechanism to capture more complex and meaningful alignment patterns~\cite{sun2023datransunet}. In this implementation, $\text{PAM}(\F^{\rm enc})$ performs the following operations: given a feature map $\F^{\rm enc} \in \mathbb{R}^{c \times h \times w}$, we apply $1 \times 1$ convolutions to generate the corresponding query $\bm Q_s$, key $\bm K_s$, and value $\bm V_s$. A spatial attention map $\bm A_s \in \mathbb{R}^{hw \times hw}$, where $hw = h \times w$, is computed using scaled dot-product attention on the spatial dimension with $\bm Q_s$ and $\bm K_s$. The final output feature is then obtained as $\text{PAM}(\F^{\rm enc}) = \bm A_s \bm V_s$. Similarly, CAM applies scaled dot-product attention along the channel dimension using $\bm Q_c$ and $\bm K_c$, resulting in a channel attention map $\bm A_c \in \mathbb{R}^{c \times c}$, with the final output feature given by $\text{CAM}(\F^{\rm pam}) = \bm A_c \bm V_c$.

Feature maps of Stages 1–3 have larger spatial dimensions, we adopt a more lightweight implementation\cite{ruan2022malunet}. For PAM, we first perform average pooling along the channel dimension on each stage’s feature map $\F^{\rm enc} \in \mathbb{R}^{c \times h \times w}$, then concatenate the two results to obtain $\bm A_s \in \mathbb{R}^{2 \times h \times w}$. Next, we process the concatenated result with the following operation to generate the spatial attention map for each stage: $\text{PAM}(\F^{\rm enc}) = \sigma(\text{conv}(\bm A_s)) \odot \F^{\rm enc}$, where $\sigma$ represents the sigmoid activation function, $\odot$ denotes element-wise multiplication, and $\text{conv}(\cdot)$ is a shared dilated convolution operation (dilation rate = 3, kernel size = 7). For CAM, we apply global average pooling to the feature maps $\F^{\rm pam}$ from all stages ($i = 1, \dots, 6$), then concatenate all pooled results along the channel dimension to obtain $\bm A_c$. A 1D convolution (kernel size = 3) is then applied to $\bm A_c$, followed by a fully connected layer with sigmoid activation to compute the channel attention weights. The final output is computed as $\text{CAM}(\F^{\rm pam}) = \sigma(\text{linear}(\text{conv}(\bm A_c)))$, where $\text{linear}(\cdot)$ refers to a fully connected layer.

\section{Experiments}
\subsection{Experimental Setup.}
\begin{figure}[!t]
\centering
\includegraphics[width=0.48\textwidth]{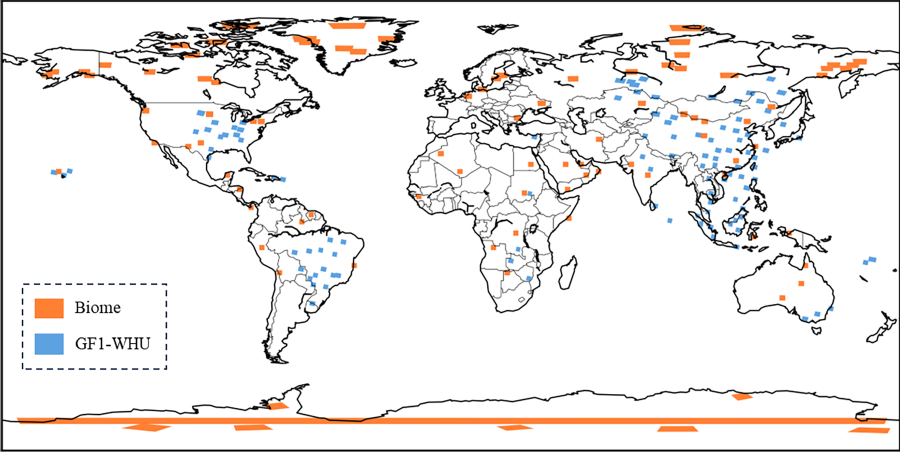}
\caption{The sampling range distribution of the Bomie and GF1-WHU datasets. Both datasets have a relatively uniform global distribution and a large sampling scale, representing global land cover and cloud conditions as comprehensively. These characteristics make them suitable for use as training and testing datasets for cloud detection tasks.}
\label{map}
\end{figure}

\subsubsection{Datesets.} As shown in Figure~\ref{map}, we employ two public datasets: Biome~\cite{foga2017cloud} and GF1-WHU~\cite{li2017multi}. Both are selected for their broad geographic diversity and global representativeness.

Biome~\cite{foga2017cloud} originates from Landsat-8 and comprises 96 high-resolution multispectral remote sensing images, each with a spatial resolution of $8,000 \times 8,000$ pixels. The dataset is evenly distributed across eight representative land cover types: barren land, forest, grassland/crops, shrubland, snow/ice, urban areas, water bodies, and wetlands, with each category containing 12 images. This ensures a balanced representation of diverse surface conditions.

The GF1-WHU~\cite{li2017multi} dataset is derived from remote sensing images captured by the GaoFen-1 satellite. Its imaging system captures four multispectral bands covering the visible to near-infrared spectral range. The dataset consists of 108 scenes, each with a spatial resolution of $16,000 \times 16,000$ pixels, including forest, barren land, snow/ice, water bodies, wetlands, and urban areas. These scenes represent a diverse range of global land cover types and varying cloud conditions.

For inputs, we use four commonly available spectral bands, red, green, blue, and near-infrared (NIR), to maintain broad compatibility with various satellite sensors. NIR is effective for cloud detection due to its high reflectance for clouds and strong contrast with vegetation, which facilitates the distinction between cloud and non-cloud regions. These four bands also allow for the derivation of widely used spectral indices such as NDVI (Normalized Difference Vegetation Index), NDWI (Normalized Difference Water Index), and EVI (Enhanced Vegetation Index). Although we do not explicitly compute these indices, CD-Mamba can implicitly learn similar spectral relationships in a data-driven manner through neural network training.

\subsubsection{Implementation Details.}
We train CD-Mamba on Biome using four-fold cross-validation and further evaluate its generalization capability on the independent GF1-WHU dataset. The 96 Biome images are divided into four folds, each containing 24 images evenly distributed across eight land cover types, with three images per type in each fold. In each cross-validation round, three folds are used for training and the remaining one for validation, ensuring that all images are used for evaluation. To further assess generalization, we select 10 large images from the GF1-WHU dataset, which cover diverse terrain types and exhibit broad spatial variation.

For Biome, each image is cropped into non-overlapping blocks of size $384 \times 384$, resulting in a total of 30,609 small images. The same cropping strategy is applied to the GF1-WHU images for consistency.

During training, CD-Mamba's parameters are initialized randomly.
We employ the AdamW optimizer~\cite{adamw} with a training schedule of 80 epochs and a batch size of 8. The learning rate is controlled using a cosine annealing scheduler~\cite{loshchilov2017sgdr}, starting from an initial value of $1 \times 10^{-3}$  with a weight decay of $1 \times 10^{-2}$ , and the minimum learning rate is set to $1 \times 10^{-5}$. The balance coefficient $\gamma$ set to 1 to equally weight the two terms of the loss.

For evaluation, we adopt several widely used metrics, including mean Intersection over Union (mIoU), $F_1$ score, and accuracy (ACC), to provide a comprehensive assessment.
\subsection{Experimental Results.}
\textbf{Comparison.} We present a performance comparison of CD-Mamba with a range of baselines based on CNNs, VITs, and vision Mamba variants. These baselines cover current state-of-the-art methods, including U-Net~\cite{ronneberger2015u}, CDNet~\cite{yang2019cdnet}, HR-cloud-Net~\cite{10.1117}, Swin-Unet~\cite{cao2022swin}, Attention Swin U-Net~\cite{aghdam2023attention}, VM-Unet~\cite{ruan2024vm}, RS-Mamba~\cite{zhao2024rs}, and UltraLight VM-UNet~\cite{wu2024ultralight}. To ensure fairness and consistency in the evaluation, we fully train all baselines on Biome using the same training strategy and parameter settings.

The cloud detection performance of different models on the Biome dataset is summarized in Table~\ref{tlb1}. In addition to these metrics, Table~\ref{cost} presents a comparison of computational complexity and inference efficiency, including FLOPs, parameter counts, and inference time, demonstrating the lightweight design and competitive speed of CD-Mamba. The results show that CD-Mamba maintains a small number of parameters while demonstrating performance comparable to or even superior to that of mainstream models in three key evaluation metrics: mIoU (87.03±2.64), $F_1$ (93.04±1.52), and ACC (94.27±0.69). It outperforms the second best model, RS-Mamba, with an average improvement of $0.42\%$  in mIoU, $0.24\%$ in $F_1$, and $0.23\%$ in ACC.

Further analysis reveals that CD-Mamba exhibits the smallest standard deviation across four different splits of Biome, which fully demonstrates its high stability and consistency under different scenarios and data splits. The model accurately identifies and segments cloud regions in various scenarios, reinforcing its reliability for real-world cloud detection applications in remote sensing.
\begin{table*}[!t]
\centering
\caption{Comparison experimental results on Biome.}\label{tlb1}
\begin{tabular*}{0.98\textwidth}{@{\extracolsep{\fill}\,}l|ccc}
\toprule
Method & mIoU &  $F_1$&  ACC
\\
\midrule
U-Net &
75.54$\pm$8.92&
85.73$\pm$5.80&
87.62$\pm$5.56
\\
CDNet &
77.59$\pm$7.66&
88.74$\pm$4.64&
90.26$\pm$2.92
\\
HR-cloud-Net &
79.94$\pm$7.24&
88.67$\pm$4.56&
90.30$\pm$3.39
\\
Swin-Unet &
86.53$\pm$3.03 &
92.75$\pm$1.66&
94.01$\pm$1.23
\\

VM-Unet &
84.98$\pm$3.79&
91.84$\pm$2.00&
93.05$\pm$2.13
\\
RS-Mamba &
86.61$\pm$3.13&
92.80$\pm$1.63&
94.04$\pm$0.90
\\
UltraLight VM-UNet &
86.79$\pm$3.72&
92.75$\pm$2.15&
93.98$\pm$1.03
\\
\midrule
CD-Mamba&
87.03$\pm$2.64&
93.04$\pm$1.53&
94.27$\pm$0.69
\\
\bottomrule
\end{tabular*}
\end{table*}

\begin{table*}[!t]
\centering
\caption{FLOPs, parameter count, and inference time comparison of cloud detection models.}\label{cost}
{\begin{tabular*}{0.98\textwidth}{@{\extracolsep{\fill}\,}l|ccc}
\toprule
Method & FLOPs (G) &  Params (M) &Inference (ms) \\
\midrule
U-Net &
90.50&
17.26&
273.17
\\
CDNet &
145.26&
46.48&
398.67
\\
HR-cloud-Net &
111.51&
75.54&
274.43
\\
Swin-Unet &
25.63&
41.34&
149.15
\\
VM-Unet &
32.11&
34.62&
51.33
\\
RS-Mamba &
36.81&
40.73&
 63.39
\\
UltraLight VM-UNet &
0.25&
0.04&
13.98
\\
\midrule
CD-Mamba&
2.12&
0.05&
38.97\\
\bottomrule
\end{tabular*}}
\end{table*}

\begin{figure*}[!t]
\begin{center}
\includegraphics[width=0.96\textwidth]{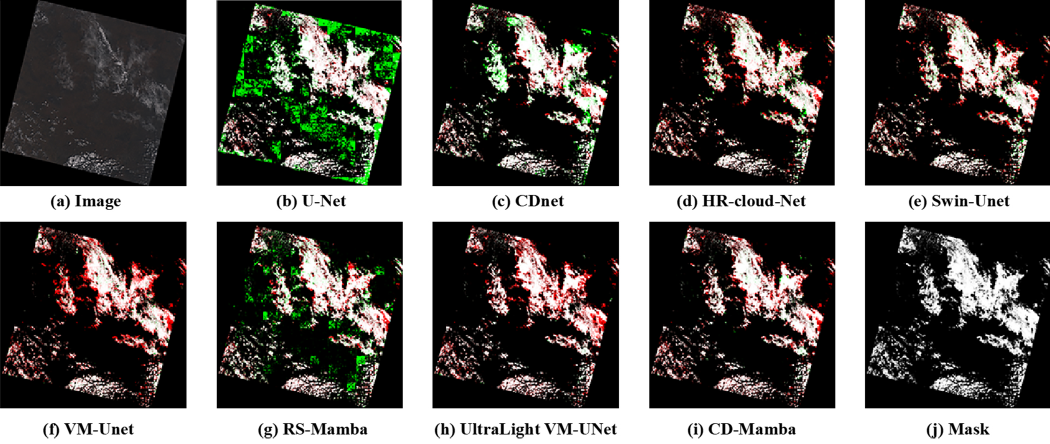}
\caption{Visual comparison between CD-Mamba and other methods on a grass/crops scene from Biome.}
\label{Biome1}
\end{center}
\end{figure*}

\begin{figure*}[!t]
\begin{center}
\includegraphics[width=0.96\textwidth]{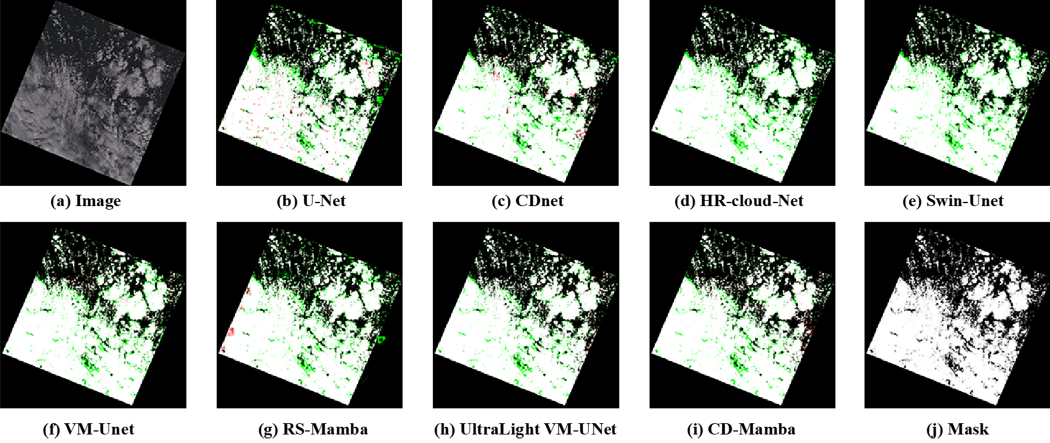}
\caption{Visual comparison between CD-Mamba and other methods on a wetland scene from Biome.}
\label{Biome2}
\end{center}
\end{figure*}

\begin{figure*}[!t]
\begin{center}
\includegraphics[width=0.96\textwidth]{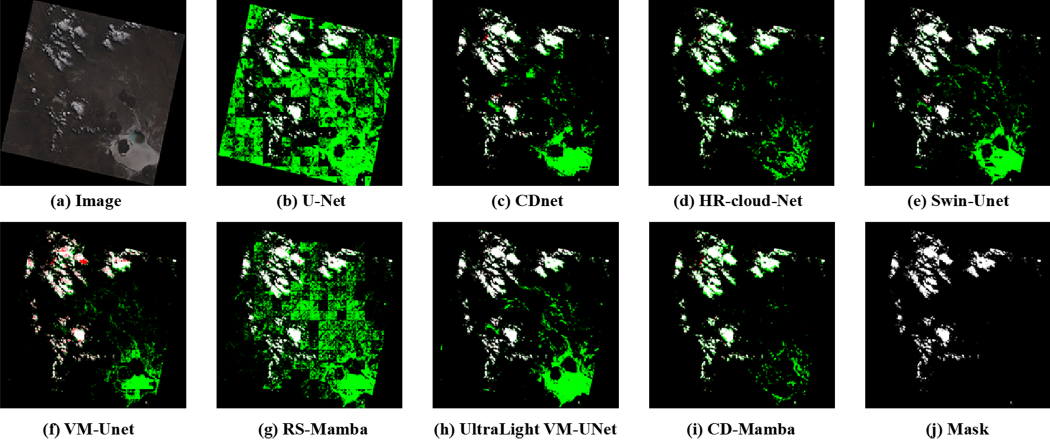}
\caption{Visual comparison between CD-Mamba and other methods on a bush scene from Biome.}
\label{Biome3}
\end{center}
\end{figure*}

Qualitative comparison of CD-Mamba and competing methods on three representative scenes from Biome is shown in Figures~\ref{Biome1}, \ref{Biome2}, and \ref{Biome3}, where red markers denote missed detections (i.e., unrecognized clouds) and green markers indicate false positives (i.e., non-cloud areas misclassified as clouds). CD-Mamba achieves better detection results with a lower error rate than other advanced networks. Specifically, Figure~\ref{Biome1} depicts a grass/crop scene, demonstrating that CD-Mamba can successfully identify most cloud-covered areas with minimal omissions. Figures~\ref{Biome2} and \ref{Biome3} show wetland and bush scenes, respectively, where CD-Mamba correctly detects most clouds compared to other models.

\begin{figure*}[!t]
\begin{center}
\includegraphics[width=0.96\textwidth]{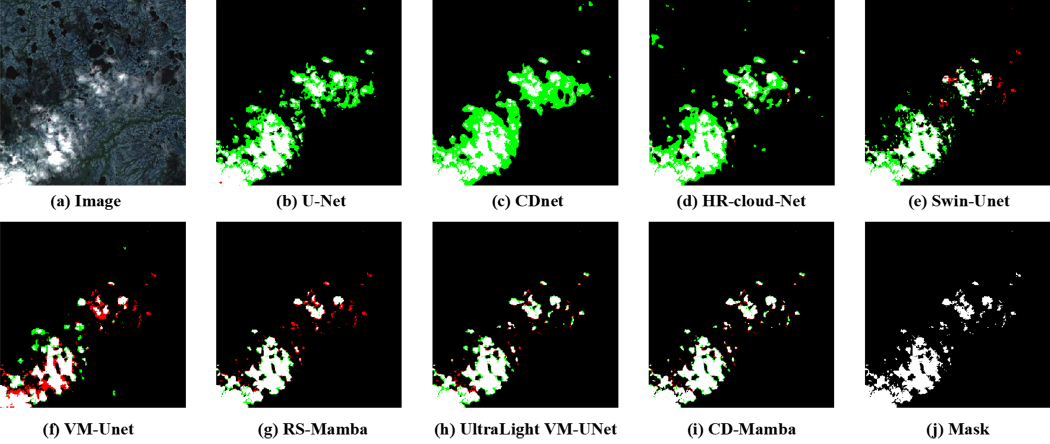}
\caption{Visual comparison between CD-Mamba and other methods in the night scene from Biome.}
\label{Biome4}
\end{center}
\end{figure*}

\begin{figure*}[!t]
\begin{center}
\includegraphics[width=0.96\textwidth]{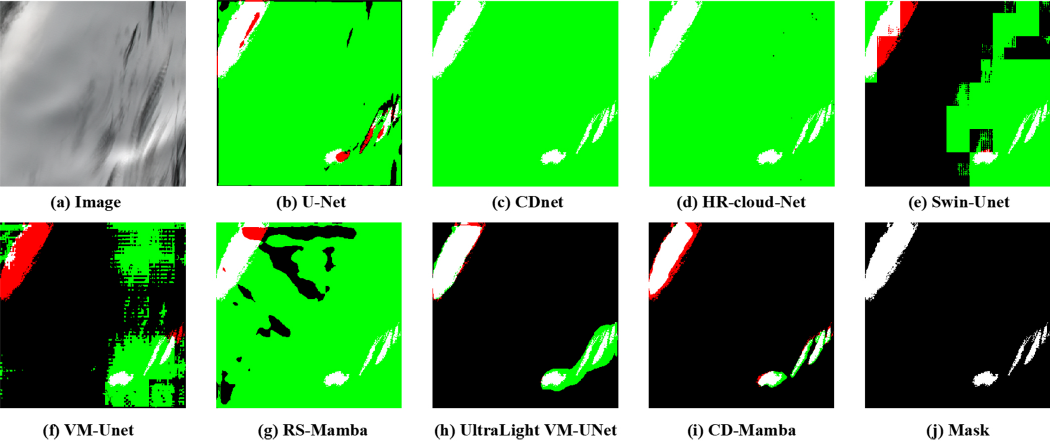}
\caption{Visual comparison between CD-Mamba and other methods in the snow scene from Biome.}
\label{Biome5}
\end{center}
\end{figure*}

Figures~\ref{Biome4} and \ref{Biome5} respectively demonstrate the performance of CD-Mamba and comparative methods in two complex scenarios: in the night scene (Figure~\ref{Biome4}), CD-Mamba shows significantly stronger recognition ability for dark-toned clouds, effectively improving the accuracy of boundary detection; while in the snowy scene (Figure~\ref{Biome5}), the model can significantly reduce the false detection rate of snow, thereby optimizing the target detection accuracy.

This effect is due to the fact that CD-Mamba effectively analyzes the spatial characteristics of clouds in remote sensing images under various scenarios through advanced deep learning techniques. Its Mamba structure is able to capture more global and fine-grained observed spatial features of clouds with different spatial structures and distribution characteristics. This capability enables CD-Mamba to achieve consistently superior performance in a variety of complex and dynamic remote sensing scenarios, providing strong support for the accurate identification of cloud regions.


Subsequently, we test a full performance evaluation of these models on GF1-WHU dataset using uniform evaluation metrics and parameter settings. In the generalization experiment on GF1-WHU, we use only one of the four training-testing partitions and conduct a comparative evaluation based on the best saved model. As shown in Table~\ref{5}, CD-Mamba achieves good results across three key evaluation metrics: mIoU (65.346), $F_1$ (79.362), and ACC (90.076). Compared to the second-best model, RS-Mamba, CD-Mamba outperforms it by an average of $2.51\%$ in mIoU, $2.19\%$ in $F_1$, and $1.58\%$ in ACC. CD-Mamba demonstrates a clear advantage and strong competitiveness in the generalization test, with the fewest errors in both false negatives and false positives. This finding validates the efficiency and robustness of the CD-Mamba model.

\begin{figure*}[!t]
\begin{center}
\includegraphics[width=0.96\textwidth]{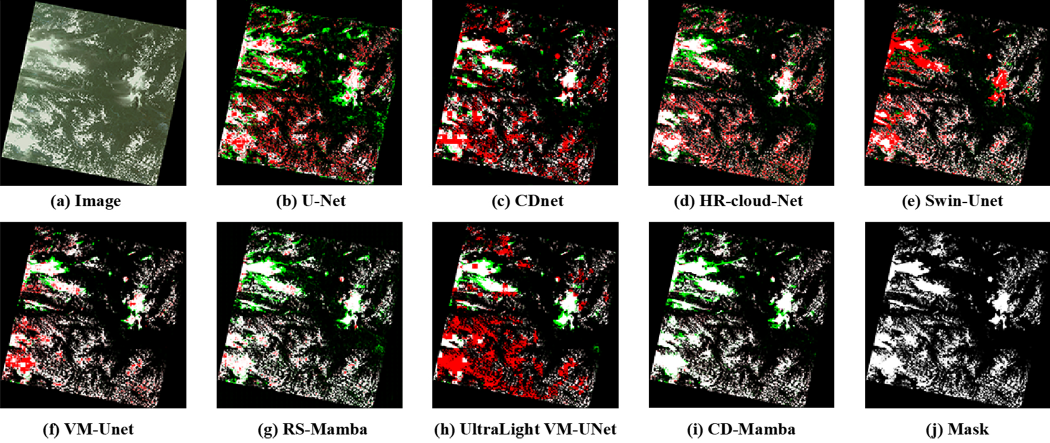}
\caption{Visual comparison between CD-Mamba and other methods on a grass/crops scene from GF1-WHU.}
\label{whu1}
\end{center}
\end{figure*}

\begin{figure*}[!t]
\begin{center}
\includegraphics[width=0.96\textwidth]{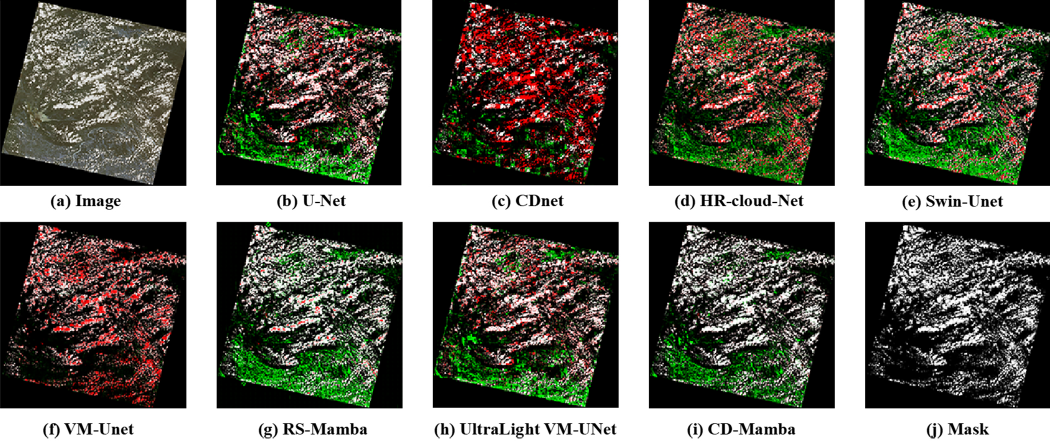}
\caption{Visual comparison between CD-Mamba and other methods on a plateau scene from GF1-WHU.}
\label{whu2}
\end{center}
\end{figure*}

\begin{figure*}[!t]
\begin{center}
\includegraphics[width=0.96\textwidth]{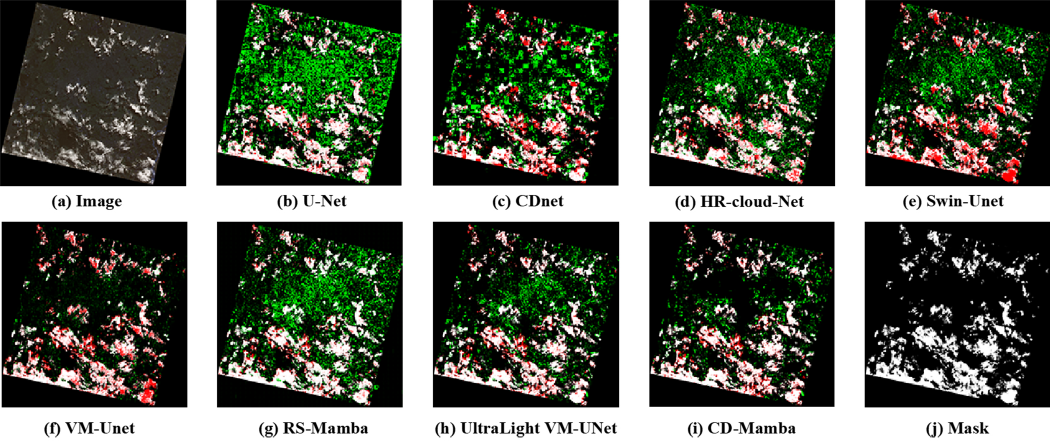}
\caption{Visual comparison between CD-Mamba and other methods on a sparse-cloud scene from GF1-WHU.}
\label{whu3}
\end{center}
\end{figure*}

We present the result of GF1-WHU dataset in Figures~\ref{whu1}, \ref{whu2}, and \ref{whu3}. Different satellites have slightly varying spectral band divisions, and the manual labeling of different datasets inevitably introduces errors.
Our approach involves processing data from various satellites to investigate whether training transfers between them can yield better cloud detection outcomes.
CD-Mamba demonstrates superior detection performance during these transfers. Figure~\ref{whu1} depicts a grassland scene, where the environment does not significantly hinder cloud detection; here, CD-Mamba exhibits enhanced recognition capabilities. Figure~\ref{whu2} showcases a plateau scene, characterized by high surface brightness, presenting a greater challenge for transfer datasets. In this scenario, CD-Mamba outperforms other networks in terms of recognition accuracy. Figure~\ref{whu3} illustrates a scenario with sparse clouds, where CD-Mamba effectively identifies cloud regions while minimizing misclassifications of similar areas. In summary, CD-Mamba demonstrates adaptability to different satellites, provides insights specifically tailored to cloud texture information, and achieves detection compatibility across various types of satellite data.

\begin{table}[!t]
\centering
\caption{Generalization Experiment on the GF1-WHU Dataset.}\label{5}
\begin{tabular*}{0.96\linewidth}{@{\extracolsep{\fill}\,}l|ccc}
\toprule
Method & mIoU &  $F_1$&    ACC\\
\midrule
U-Net &
37.184& 54.211& 74.312\\
CDNet &
42.884 &60.026 &80.883\\
HR-cloud-Net &
46.848 &63.085 &84.400\\
Swin-Unet&
29.421& 45.466& 80.195\\
RS-Mamba &
62.834& 77.175& 88.495\\
VMUNet&
61.525&76.180 &90.013 \\
UltraLight VM-UNet&
57.818& 73.271& 88.628\\
\midrule
CD-Mamba&
65.346& 79.362& 90.076\\
\bottomrule
\end{tabular*}
\end{table}


\textbf{Ablation Study.}
We design a series of ablation experiments to systematically analyze and evaluate the contribution of different modules in CD-Mamba for the cloud detection task.
Specifically, we examine the importance of the Cloud-SMB and DA-Block by removing them from the model. Since Cloud-SMB is inherently integrated into Mamba, it cannot be removed directly. To simulate a scenario without the Cloud-SMB, we replace it with the original Mamba module and use this setup in the following experiments.
Next, we evaluate the impact of the DA-Block on cloud detection accuracy by disabling the Cloud-SMB. We also construct models that retain only Cloud-SMB, removing the DA-Block, to analyze the role of the Cloud-SMB in the cloud detection task.

\begin{figure}[!t]
\centering
\includegraphics[width=0.46\textwidth]{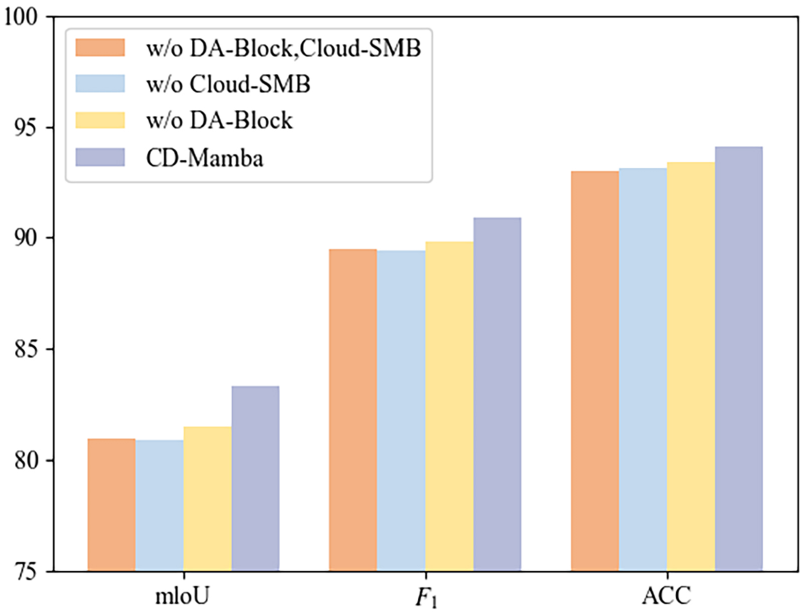}
\caption{Ablation study on the contribution of individual modules to CD-Mamba.}
\label{ab1}
\end{figure}

\begin{figure}
    \centering
    \includegraphics[width=0.90\linewidth]{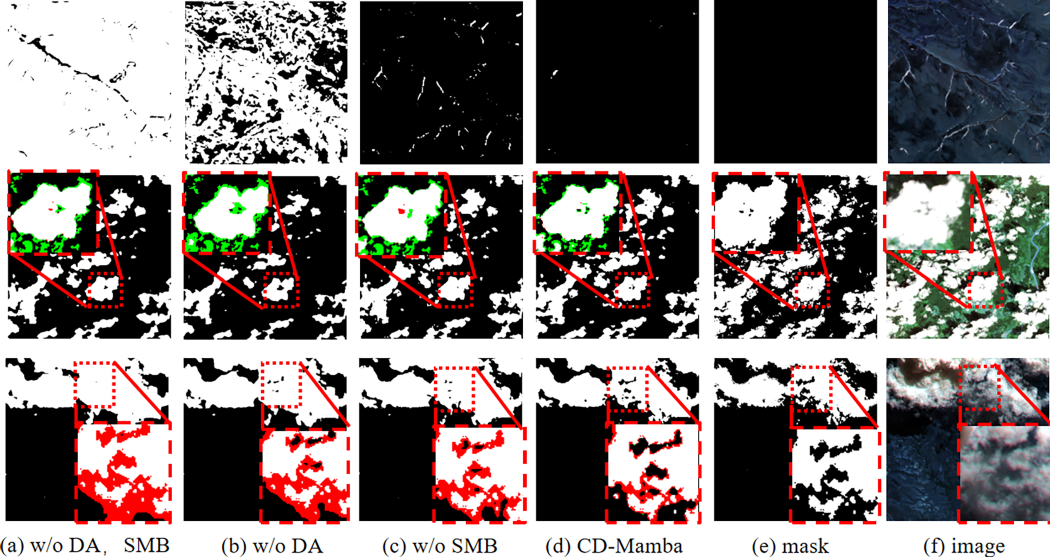}
    \caption{Visual comparison of the ablation study results on Biome.}
    \label{ab-Biome}
\end{figure}

The results of this ablation experiment are shown in Figure~\ref{ab1}, which show that eliminating both DA-Block and Cloud-SMB leads to a decrease in model performance. This indicates that each component plays a crucial role in the overall model. To further show the contribution of the different components in addressing the particular problem, we visualize the experimental results of ablation on Biome as shown in Figure~\ref{ab-Biome}. The first row of Figure~\ref{ab-Biome} shows the results of the ablation experiments under the interference of a specific terrain (rain and snow coexistence). After removing the DA-Block, the cloud misdetection problem increases significantly. This indicates that the attention mechanism and skip connection design of DA-Block enable more refined extraction and aggregation on the input data, especially under complex scenes when dealing with snow and ice or high brightness. These capabilities allow the model to better distinguish clouds from visually similar backgrounds, thereby reducing false detections and improving overall performance. In the second and third rows of results in Figure~\ref{ab-Biome}, a comparative analysis shows that Cloud-SMB performs well in spatial feature extraction. Specifically, we zoom in on cloud regions exhibiting boundary ambiguity features and find that Cloud-SMB can effectively supplement these detailed characteristics, ensuring complete and accurate extraction of cloud features. This mechanism significantly mitigates boundary ambiguity issues and enhances cloud detection accuracy. Experimental results demonstrate that both the Cloud-SMB and DA-Block modules play pivotal roles in CD-Mamba, with their effectiveness being thoroughly validated.

\begin{figure}
    \centering
    \includegraphics[width=0.90\linewidth]{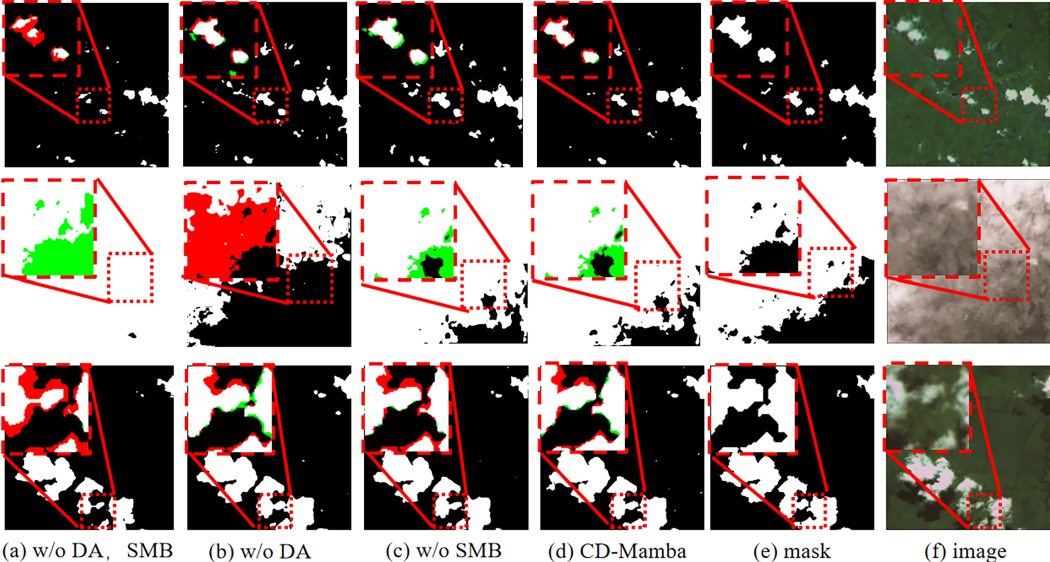}
    \caption{Visual comparison of the ablation study results on the GF1-WHU dataset.}
    \label{ab-whu}
\end{figure}

Similarly, we conduct ablation experiments on the GF1-WHU dataset and the results are shown in Figure~\ref{ab-whu}. Compared to the results with different components removed, CD-Mamba identifies the cloud boundary more accurately, which also implies a better generalization. The experiments further validate the effectiveness of our network in coping with the natural boundary ambiguity of clouds and terrain confusion interference.


To thoroughly analyze the performance of convolutional layers and Mamba modules in deep neural networks, and rigorously demonstrate the rationality of the integration strategy of convolutional modules and Cloud-SMB modules in the CD-Mamba hybrid model, we conduct a series of systematic ablation experiments. This experimental framework is built upon the symmetric U-Net structure, and by carefully adjusting the configuration ratio of convolutional layers and Cloud-SMB modules, we aim to reveal the impact of different combinations on the network performance.
Specifically, we design five experimental configurations: (1) a pure convolutional network structure that serves as a baseline control; (2) a network that fully employs Cloud-SMB modules to evaluate the performance of the Cloud-SMB when it operates independently; (3) a hybrid structure that contains two convolutional layers and four Cloud-SMB layers; (4) a configuration with a balanced number of convolutional and Cloud-SMB layers (three each); and, (5) a specific configuration adopted by CD-Mamba. This series of experiments aims to fully explore the synergy between convolution and Cloud-SMB at different depths.

\begin{figure}[!t]
\centering
\includegraphics[width=0.48\textwidth]{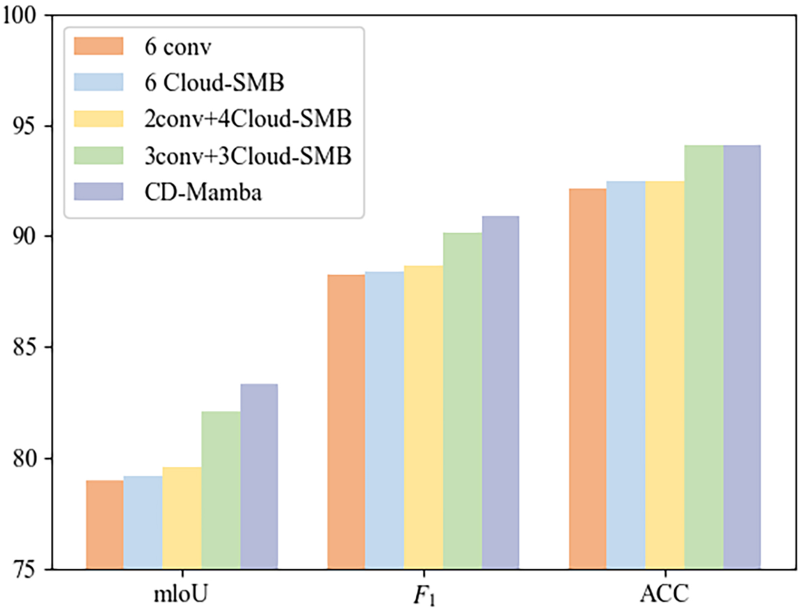}
\caption{Ablation experiments on the effect of Cloud-SMB.}
\label{ab2}
\end{figure}

As shown in Figure~\ref{ab2}, the experimental results clearly reveal several key findings. First, the pure convolutional structure significantly underperforms the network that fully employs Cloud-SMBs, which highlights the advantages of Cloud-SMBs in deep learning, especially when dealing with complex feature relationships. However, it is worth noting that in shallower network hierarchies, the convolutional layer outperforms Cloud-SMB by efficiently capturing short-term features. This observation leads to an important conclusion: while the long-range dependency capture ability of Cloud-SMBs is more effectively leveraged in deeper architectures, the convolutional layer is more adept at capturing short-term dependencies among local pixels in shallow layers, thus demonstrating higher processing efficiency. Based on the above findings, we suggest that, when constructing high-performance neural networks, the ratio of convolutional layers to Cloud-SMBs should be reasonably deployed by considering the relationship between task requirements and network depth. This strategy not only helps to improve the overall performance of the network, but also provides a theoretical foundation for designing more efficient and adaptable network architectures in the future. Nevertheless, this study is still in the preliminary exploration stage, and the optimal fusion of convolutional layers and Cloud-SMBs and their performance in different application scenarios still requires further investigation and validation.

\begin{table}[!t]
\centering
\caption{Ablation study of different terms of the loss function.}
\label{Ablation_tlb}
\begin{tabular}{p{1.1cm}p{1.1cm}|p{1.2cm}p{1.2cm}p{1.2cm}p{1.2cm}}
\toprule
$\mathcal{L}_{\rm dice}$  &$\mathcal{L}_{\rm bce}$  & mIoU &  $F_1$&  ACC\\
\midrule
\checkmark & &68.06&81.00&86.40\\
   &\checkmark&82.75& 90.56&93.82\\
\checkmark  &\checkmark&83.32 &90.90&94.07\\
\bottomrule
\end{tabular}
\label{Ablation}
\end{table}

As shown in Table~\ref{Ablation_tlb}, the ablation study evaluates the impact of different terms of the loss function on CD-Mamba performance. These results clearly demonstrate that combining $\mathcal{L}_{\rm bce}$ and $\mathcal{L}_{\rm dice}$ leads to superior performance across all key metrics. $\mathcal{L}_{\rm dice}$ alone is unstable in early training due to its sensitivity to small objects and imbalanced classes, while $\mathcal{L}_{\rm bce}$ provides stable pixel-wise supervision but lacks sensitivity to class imbalance. The combination leverages the strengths of both: $\mathcal{L}_{\rm bce}$ stabilizes optimization, while $\mathcal{L}_{\rm dice}$ enhances segmentation quality by focusing on the overlap between prediction and ground truth.

\section{Conclusion}
This paper presents CD-Mamba, a pioneering model that applies an enhanced Mamba-based architecture to cloud detection for the first time. By combining convolutional layers with an optimized Mamba structure, we design a network that fully utilizes spatial relationship information in remote sensing images and thoroughly analyzes cloud textures. This model effectively preserves detailed features while significantly narrowing the semantic gap during cloud texture extraction.
We present an innovative enhancement to the original Mamba module by introducing the Spatial Mamba Block. This new module employs a parallel processing mechanism to improve computational efficiency and utilizes a multi-directional image scanning strategy to more effectively capture complex spatial relationships within two-dimensional images. The proposed design enables CD-Mamba to efficiently model long-range spatial dependencies, which is critical for achieving high accuracy and robustness in cloud detection tasks.

Additionally, by integrating a Dual Attention Block into the skip connections, we optimize the fusion of spatial and channel features. This mechanism enhances the interaction between shallow and deep features, improving feature representation quality. Experiments indicate that this attention mechanism helps the model identify critical cloud regions more effectively while reducing redundancy, thereby boosting overall performance. We test CD-Mamba extensively on multiple remote sensing datasets, demonstrating its excellent generalization and stability across diverse data sources and complex cloud coverage scenarios. These results strongly support Mamba-based vision tasks\cite{zhu2024vision,yang2024vivim,liu2024vmamba,liu2024swin,liu2024visionmamba,Yu_2025_CVPR,maxjars2025} and provide an effective solution for the future development of remote sensing image processing technologies.
\section*{Disclosures}
The authors declare that there are no financial interests, commercial affiliations, or other potential conflicts of interest that could have influenced the objectivity of this research or the writing of this paper.
\section*{Code, Data, and Materials Availability}
The source code is available at \url{https://github.com/kunzhan/CD-Mamba}. The download links for the two datasets are provided in README.md of the GitHub repository.
\section*{Acknowledgments}
This work was supported by the Fundamental Research Funds for the Central Universities under Nos.~lzujbky-2022-ct06 and lzujbky-2024-it55, and Supercomputing Center of Lanzhou University.